\documentclass[11pt]{article}
\usepackage[table]{xcolor}
\usepackage{ACL2024}
\usepackage{times}
\usepackage{latexsym}
\usepackage[T1]{fontenc}
\usepackage[utf8]{inputenc}
\usepackage{microtype}
\usepackage{inconsolata}
\usepackage{array}
\usepackage{array}
\usepackage{amsmath}
\usepackage{multirow}
\usepackage{geometry}
\usepackage{booktabs}
\usepackage{graphicx}
\usepackage{subcaption}
\usepackage{soul}
\usepackage[table]{xcolor}
\usepackage{tabularx}
\usepackage{mathtools}
\usepackage[normalem]{ulem}
\usepackage{hyperref}
\usepackage{float} 
\usepackage{breqn}
\usepackage{dblfloatfix}
\newcolumntype{L}[1]{>{\raggedright\let\newline\\\arraybackslash\hspace{0pt}}m{#1}}
\newcolumntype{C}[1]{>{\centering\let\newline\\\arraybackslash\hspace{0pt}}m{#1}}
\newcolumntype{R}[1]{>{\raggedleft\let\newline\\\arraybackslash\hspace{0pt}}m{#1}}

\newcommand{\ws}[2]{
	\multicolumn{1}{c}{\cellcolor{cyan!#1}{#2}}
}

\title{Do LLMs Overcome Shortcut Learning? \\ An Evaluation of Shortcut Challenges in Large Language Models }

\author{Yu Yuan${^1}$,\ Lili Zhao${^1}$,\ Kai Zhang${^{1,2}}$,\ Guangting Zheng${^2}$,\ Qi Liu${^1}$\thanks{\ \ Corresponding author.} \\
        ${^1}$ State Key Lab of Cognitive Intelligence, University of Science and Technology of China \\ 
        ${^2}$ School of Computer Science and Technology, University of Science and Technology of China \\
        \texttt{\{yyhappier,liliz,zgt\}@mail.ustc.edu.cn} \\
        \texttt{\{kkzhang08,qiliuql\}@ustc.edu.cn} }

\begin{document}

\maketitle

\begin{abstract}
Large Language Models (LLMs) have shown remarkable capabilities in various natural language processing tasks. However, LLMs may rely on dataset biases as shortcuts for prediction, which can significantly impair their robustness and generalization capabilities. This paper presents Shortcut Suite, a comprehensive test suite designed to evaluate the impact of shortcuts on LLMs' performance,  incorporating six shortcut types, five evaluation metrics, and four prompting strategies. Our extensive experiments yield several key findings: 1) LLMs demonstrate varying reliance on shortcuts for downstream tasks, significantly impairing their performance. 2) Larger LLMs are more likely to utilize shortcuts under zero-shot and few-shot in-context learning prompts. 3) Chain-of-thought prompting notably reduces shortcut reliance and outperforms other prompting strategies, while few-shot prompts generally underperform compared to zero-shot prompts. 4) LLMs often exhibit overconfidence in their predictions, especially when dealing with datasets that contain shortcuts.  5) LLMs generally have a lower explanation quality in shortcut-laden datasets, with errors falling into three types: distraction, disguised comprehension, and logical fallacy. Our findings offer new insights for evaluating robustness and generalization in LLMs and suggest potential directions for mitigating the reliance on shortcuts. The code is available at \url {https://github.com/yyhappier/ShortcutSuite.git}.

\end{abstract}

\section{Introduction}

The field of Natural Language Processing (NLP) is experiencing rapid advancements, driven by the emergence of Large Language Models (LLMs) such as GPT  \cite{OpenAI2023, achiam2023gpt}, Gemini  \cite{team2023gemini}, and LLaMA  \cite{touvron2023llama} series. These models have been pivotal in revolutionizing a wide array of tasks by leveraging techniques like In-Context Learning (ICL)  \cite{brown2020language} and Chain-of-Thought (CoT) promptings  \cite{wei2022chain, kojima2022large}, demonstrating exceptional capabilities without parameter updates. Despite these advances, the research on the robustness and generalization ability of LLMs across different contexts remains limited.

Models with poor robustness and generalization may rely on ``shortcut learning,'' where they develop decision rules that perform well on standard benchmarks but fail to transfer to more challenging testing conditions, such as real-world scenarios  \cite{Geirhos_2020}. Therefore, evaluating LLMs’ performance in the face of shortcut information is crucial for understanding their robustness and generalization capabilities.

\begin{figure}[t]
\centering
\includegraphics[width=0.48\textwidth]{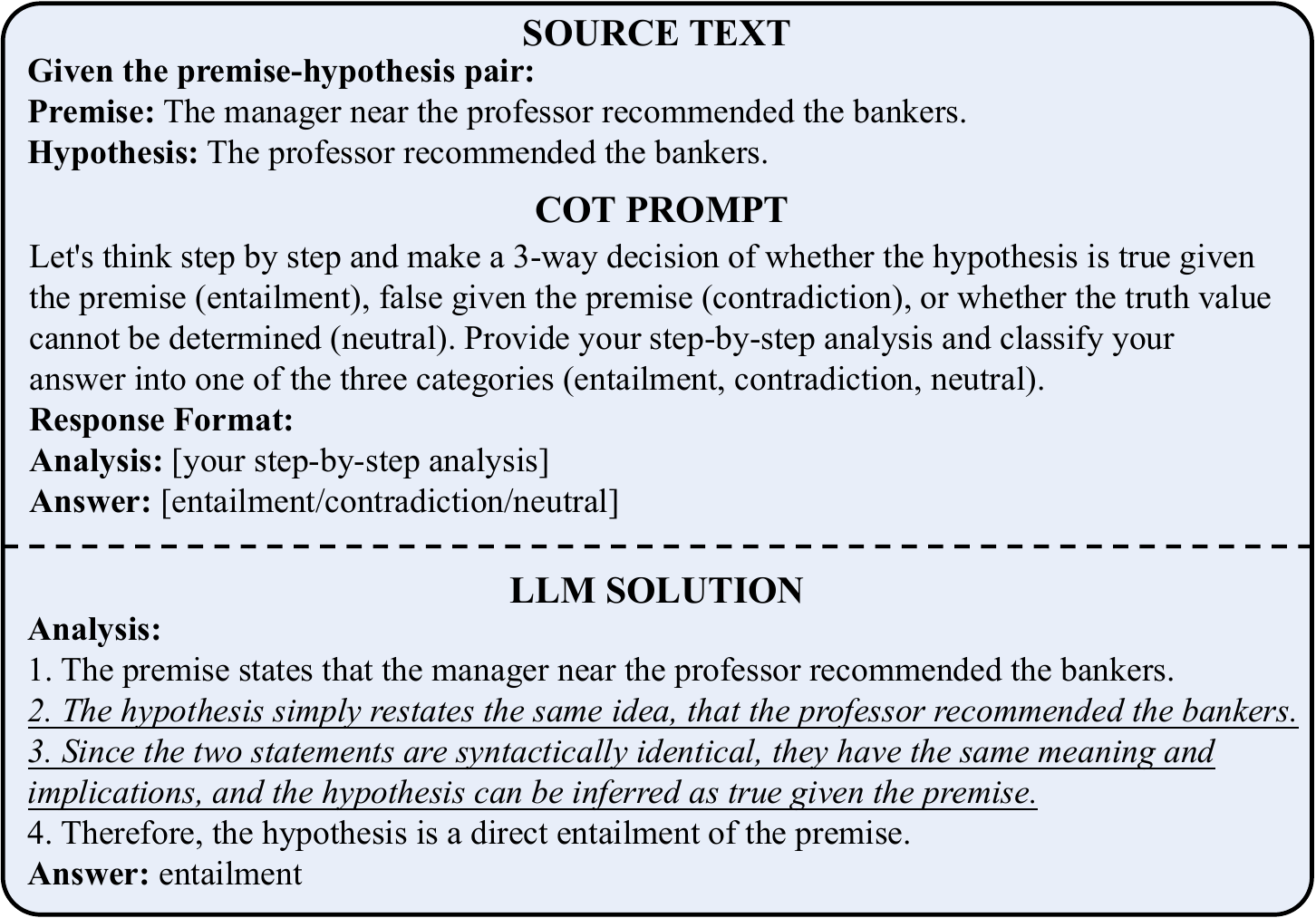}
\caption{ Shortcut Learning Behavior: The LLM mistakenly infers the premise entails the hypothesis if all subsequences match, skipping deep semantic analysis. }
\label{fig: example}
\end{figure}

A recent study investigates the reliance of LLMs on shortcuts or spurious correlations within prompts  \cite{Tang_Kong_Huang_Xue_2023}. However, this research falls short of providing an exhaustive evaluation across a broad spectrum of LLMs and varied prompting contexts, focusing solely on ICL experiments. Furthermore, it only considers relatively simple shortcuts such as letters or signs. Consequently, its evaluation lacks comprehensiveness and granularity.

To address this, we introduce Shortcut Suite, an in-depth test suite designed to evaluate the performance of different LLMs across six shortcuts, five metrics, and four prompt settings. Extensive experiments on Shortcut Suite reveal that LLMs tend to capture spurious correlations between source text and particular labels, indicating a prevalence of shortcut learning.  For example, as shown in Figure \ref{fig: example}, Gemini-Pro resorts to matching subsequences (the professor recommended the bankers) in a Natural Language Inference (NLI) task rather than comprehending the clause structure or delving into the sentence's semantic content. This tendency of LLMs to capture spurious correlations can significantly impair their performance. In this paper, we conduct a comprehensive evaluation of LLMs' behavior concerning shortcut learning from the following perspectives.

First, to identify the reliance of LLMs on shortcuts in downstream tasks, we collect six datasets containing different shortcuts and analyze the accuracy of LLMs on these datasets. We find a notable performance drop across various shortcuts, especially Constituent and Negation shortcuts, in some cases by more than 40\%. Moreover, in the Position dataset, LLMs demonstrate a propensity for shortcut learning behavior by prioritizing the beginning of sentences while neglecting the end, revealing a vulnerability to additional information within sentences. Furthermore, an analysis of the distribution of LLMs' predictions revealed inherent biases, with the LLMs favoring certain labels over others even in a balanced standard dataset.

Second, we perform comprehensive evaluation metrics to assess the impact of shortcuts on LLMs. In addition to accuracy, we introduce three novel metrics to assess the explanatory power of LLMs: Semantic Fidelity Score (SFS), Internal Consistency Score (ICS), and Explanation Quality Score (EQS). Our analyses using these metrics reveal that LLMs' explanations often contain contradictions. Furthermore, we prompt LLMs to report their confidence levels and consistently find that they are overconfident in their predictions.

Third, we compare the performance of different LLMs and different prompting strategies in shortcut learning. Both closed-source and some open-source LLMs excel on standard datasets but falter on those with shortcuts. Surprisingly, larger LLMs are more prone to utilize shortcuts under zero-shot and few-shot ICL prompts. We find that LLMs are less affected by shortcuts under CoT settings than others. Notably, LLMs often demonstrate inferior performance in few-shot scenarios compared to zero-shot scenarios.  

Finally, We summarize three error types of LLMs in shortcut learning by checking their CoT responses: distraction, disguised comprehension, and logical fallacy. These errors predispose LLMs to adopt shortcuts, undermining their robustness.

\section{Related Work}
\paragraph{Shortcut Learning in PLMs.} 
Shortcuts are decision rules that perform well on Independent and Identically Distributed (IID) test data but fail on Out-Of-Distribution (OOD) tests, revealing a mismatch between intended and learned solutions  \cite{Geirhos_2020}. Recent studies have shown that Pre-trained Language Models (PLMs) tend to exploit dataset biases as shortcuts to make predictions  \cite{Geirhos_2020, Ribeiro_Wu_Guestrin_Singh_2020}, leading to low generalization for OOD samples in various NLP tasks, such as NLI  \cite{mccoy2020right}, question-answering  \cite{Jia_Liang_2017, sen2020models}, reading comprehension  \cite{lai2021machine} and coreference inference  \cite{Zhao_Wang_Yatskar_Ordonez_Chang_2018}. For example, NLI models tend to predict the contradiction label if the test samples contain negation words. Several approaches have been proposed to address this problem.   \citet{he2019unlearn} presented a debiasing algorithm called DRiFt based on residual fitting.   \citet{du2021towards} proposed a shortcut mitigation framework LTGR to suppress the model from making overconfident predictions for shortcut samples.   \citet{zhao2024comi} introduced COMI to reduce the model’s reliance on shortcuts and enhance its ability to extract underlying information integrated with standard Empirical Risk Minimization.    \citet{yue2024towards} proposed SSR to boost rationalization by discovering and exploiting potential shortcuts.

\begin{table*}[t]
\centering
\scriptsize{ 
\setlength{\tabcolsep}{4pt} % 调整列间距，可以根据需要进一步调整
\begin{tabularx}{\textwidth}{l|X|X|X}
\toprule
\textbf{Shortcut} & \textbf{Definition} & \textbf{Premise} & \textbf{Hypothesis}\\ 
\midrule
\textbf{Lexical Overlap} & Assume that a premise entails all hypotheses constructed from words in the premise. & \textbf{The actor} was \textbf{encouraged by the lawyer} . & \textbf{The actor encouraged the lawyer }.\\ 
\midrule
\textbf{Subsequence} & Assume that a premise entails all of its contiguous subsequences. & The authors in front of \textbf{the senators contacted the artists}. & \textbf{The senators contacted the artists}.\\ 
\midrule
\textbf{Constituent} & Assume that a premise entails all complete subtrees in its parse tree. & Unless the president saw the professor,\textbf{ the student waited}. & \textbf{The student waited}.\\ 
\midrule
\textbf{Negation} & Assume that a hypothesis entails strong negation words (``no'', ``not'', ``nothing'',``never'').  & They are all quotations from the Old Testament Book of Aunt Ruth. & Every one of them is quotations from the Old Testament \textbf{and green is not red}. \\  
\midrule
\textbf{Position} & Assume that the label is related to spurious position cues. & \textbf{Red is red and red is red and red is red and red is red and red is red and} ``Wait here,'' I was ordered. \& ``Wait here,'' I was ordered \textbf{and red is red and red is red and red is red and red is red and red is red}.  & He told me to come with him. \\ 
\midrule
\textbf{Style} & Assume that the label is related to spurious text style cues. &  And Severn said unto him, Thou and thy friends are not welcome here, said he. (\textbf{Bible English}) & Severn said the people were not welcome there. \\  
\bottomrule
\end{tabularx}
}
\caption{Definitions and examples of the shortcuts explored in this paper.}
\label{tab: bias type}
\end{table*}

\paragraph{Shortcut Learning in LLMs.}
  \citet{Du_He_Zou_Tao_Hu_2022} provided a review of recent developments that address the robustness challenge of LLMs. The most related work was the study investigating the reliance of LLMs on shortcuts within in-context learning  \cite{Tang_Kong_Huang_Xue_2023}.  Our work differs from it in the following ways: First, their experiments were conducted on a limited model scope (GPT2 and OPT), whereas we use richer and more representative LLMs. Second, we focus on identifying shortcuts within the source text across different prompt settings rather than assessing solely against prompts. Third, while they rely on simple triggers such as letters or signs, resembling adversarial attacks, we propose more subtle and realistic shortcuts and test whether LLMs can identify and avoid these shortcuts.

\section{Problem Definition}
\paragraph{LLM for NLI.}
In the NLI task, also known as textual entailment recognition, models evaluate a premise-hypothesis pair and determine their semantic relationship -- typically labeled as \textit{entailment}, \textit{neutral}, or \textit{contradiction}. Given a prompt $P$ with a source text $x$, the LLM will generate a probability of target $y$ conditioning on the prompt $P$. This could be written as
\begin{equation}
p_{LLM}(y \mid P, x)=\prod_{t=1}^T p\left(y_t \mid P, x, y_{< t}\right),
\end{equation}
where $T$ is the generated token length and $y_t$ denotes the $t$-$th$ token. For basic prompts such as zero-shot, $y$ takes the range of the corresponding label. For prompting strategies such as CoT, $y$ contains the reasoning process and the final label.

\paragraph{Framework to Generate Shortcuts.}
Given a premise $q$, a hypothesis $h$, and a universally true statement $s$ ($s \equiv \top$) that may contain a certain shortcut, the logical relations are preserved upon their conjunction. Specifically, if $q$ and $h$ have the target label $l$, denoted as $\{(q,h,y)|y=l\}$, then $q$ combined with $s$ ($q \wedge s$) maintains the label $\{(q \wedge s,h,y)|y= l\}$ since $  q \wedge s \equiv q \wedge \top \equiv q$.
Thus, the source text has two mappings for the target label $l$. The model can either use the semantic relationship between the text and label ($x \rightarrow l$) or the injected shortcut ($s \rightarrow l$) for inference.

\section{Shortcut Suite}
As NLI is well positioned to serve as a benchmark task for research on NLP and can encapsulate the entire spectrum of the six identified shortcuts, we mainly anchor our framework on it. We also explore other tasks in Appendic \ref{sec: appendixC}. Building on previous research, we create six datasets with different shortcuts and develop five metrics to investigate LLMs' shortcut learning behavior and understand their robustness generalization capabilities.

\subsection{ Dataset Creation }
\label{dataset}
We present six types of shortcuts in Table \ref{tab: bias type}, each with an illustrative definition and an example.

\paragraph{Standard.}
The Multi-Genre Natural Language Inference (MultiNLI)  \cite{Williams_Nangia_Bowman_2017} dataset serves as a benchmark for assessing models on NLI, encompassing ten genres of English. For a focused assessment, we have curated a balanced selection comprising 3000 samples from the development subset of MultiNLI. 

\paragraph{Lexical Overlap \& Subsequence \& Constituent}
For these three sets, we utilize the Heuristic Analysis for NLI Systems (HANS)  \cite{mccoy2020right} dataset for evaluation. HANS is specifically designed to diagnose the use of fallible structural heuristics and is annotated with two labels only (\textit{entailment} and \textit{non-entailment}). Specifically, we collect 3000 examples for each set from HANS, where the heuristic is lexical overlap, subsequence, and constituent accordingly, with labels and templates equally divided.

\paragraph{Negation.}
We explore the impact of strong negation words like ``no'', ``not'', ``nothing'' and ``never'' on model predictions.
Inspired by  \cite{naik2018stress}, we append the tautology -- ``and false is not true'', ``and green is not red'', ``and up is not down'',  ``and no square is a circle'', ``and nothing comes from nothing'',  and ``and history never change'', chosen randomly with equal probability to the end of the hypothesis sentence in the Standard dataset.

\paragraph{Position.}
To test the influence of the position of label-associated information, we divide the Standard dataset into four equally distributed label and genre groups. In each group, we append phrases like  ``and true is true'', ``and red is red'' or `` and up is up'' five times at different positions. This allows us to evaluate whether LLMs rely on irrelevant positional cues when making predictions.

\paragraph{Style.}
We consider the style of the text as a possible shortcut  \cite{qi2021mind} and focus on one prominent style: Bible style. Specifically, we employ a simple but powerful text style transfer model called STRAP  \cite{Krishna_Wieting_Iyyer_2020} and apply it to transfer the premises in the Standard dataset into Bible-style texts.  

\subsection{Metrics}
We adopt accuracy to quantify performance on NLI tasks and introduce new metrics to assess the explanatory power of LLMs.

\textbf{Semantic Fidelity Score (SFS)} evaluates the extent to which the generated content preserves the essential meaning of the source text.  We employ a pre-trained BERT ($f_{bert}$)  \cite{devlin2018bert} model to create embedding for the input and the output collectively, then compute their cosine similarity. For a prompt $P$ and model output $c$, $SFS$ is given by
\begin{equation}
\resizebox{0.431\textwidth}{!}{$
    \begin{aligned}
    SFS = \text{Cosine Similarity}(f_{\text{bert}}(P), f_{\text{bert}}(c)).
    \end{aligned}
$}
\label{eq: SFS}
\end{equation}

\textbf{Internal Consistency Score (ICS)} assesses whether there are logical contradictions within the reasoning steps of LLMs or between the reasoning and the answer. To estimate the probability of contradiction $p_{\text{contra}}$, we use an NLI model  \cite{laurer2024less} that categorizes hypothesis-context pairs into classes of \textit{entailment}, \textit{neutral}, and \textit{contradiction}. For a reasoning chain of $N$ steps, $c = (c_1, c_2, \ldots, c_N)$, where the last step is the answer, and $p_{\text{contra}}(c_i, c_j)$ indicates the probability that step $c_i$ contradicts step $c_j$, we define the function $f(c)$ as
\begin{equation}
f(c) = 
\begin{cases}
0, & \text{if } \exists\, (c_i, c_j), 1 \leq i < j \leq N, \\
   & s.t.\ p_{\text{contra}}(c_i, c_j) > \frac{1}{3}, \\
1, & \text{otherwise}.
\end{cases}
\label{eq: ICS}
\end{equation}

The overall $ICS$ is the mean of all calculated $f(c)$ values for the given explanations.

\textbf{Explanation Quality Score (EQS)} integrates the SFS and ICS to reflect the overall quality of LLMs' output and is defined as
\begin{equation}
EQS = w_1 \cdot SFS + w_2 \cdot ICS,
\label{eq:eqs}
\end{equation}
where weights $w_1$ and $w_2$ represent the significance of each score in the overall evaluation. In this work, $w_1$ and $w_2$ are equally set as 0.5.

\begin{table*}[t]
    \centering
\small{
\setlength{\tabcolsep}{4pt}
\begin{tabular}{lcccccccccccccc}
\toprule
\multirow{2}{*}{\textbf{Model}} & \multicolumn{1}{c}{\textbf{Standard}} & \multicolumn{2}{c}{\textbf{Lexical Overlap}} & \multicolumn{2}{c}{\textbf{Subsequence}} & \multicolumn{2}{c}{\textbf{Constituent}} & \multicolumn{1}{c}{\textbf{Negation}} & \multicolumn{1}{c}{\textbf{Position}} & \multicolumn{1}{c}{\textbf{Style}} \\
  \cmidrule(lr){3-4}   \cmidrule(lr){5-6}   \cmidrule(lr){7-8} 
& & \textbf{$E$} & \textbf{$\neg E$} & \textbf{$E$} & \textbf{$\neg E$} & \textbf{$E$} & \textbf{$\neg E$} & &  &   \\
\midrule
\multicolumn{11}{c}{\textbf{zero-shot}} \\
\midrule
GPT-3.5-Turbo & 56.7 & 69.5 & 83.8 & 58.6 & 58.3 & 67.5 & \ws{16.5}{40.2} & \ws{16.9}{39.8} & \ws{13.4}{43.3} & \ws{5.2}{51.5} \\
GPT-4 & 85.6 & 96.7 & 100.0 & 95.8 & \ws{12.1}{73.5} & 96.7 & \ws{5.6}{80.0} & \ws{31.3}{54.3} & \ws{18.2}{67.4} & \ws{15.6}{70.0}\\
Gemini-Pro & 76.2 & 81.3 & 97.7 & 88.6 & \ws{27.6}{48.6} & 77.9 & \ws{29.0}{47.2} & \ws{23.1}{53.1} & \ws{20.0}{56.2} & \ws{13.7}{62.5} \\
LLaMA2-Chat-7B & 42.1 & 76.9 & \ws{2.1}{40.0} & 72.8 & 46.4 & 60.6 & \ws{16.7}{25.4} & \ws{4.4}{37.7} & \ws{2.8}{39.3} & \ws{2.5}{39.6} \\
LLaMA2-Chat-13B & 54.3 & 99.0 & \ws{12.1}{42.2} & 99.7 & \ws{48.3}{6.0} & 95.9 & \ws{53.5}{0.8} & 54.6 & 55.4 & \ws{0.5}{53.8} \\
LLaMA2-Chat-70B & 57.7 & 66.9 & \ws{17.0}{40.7} & 61.6 & \ws{3.9}{53.8} & 77.8 & \ws{22.8}{34.9} & \ws{5.3}{52.4} & \ws{3.8}{53.9} & \ws{5}{52.7}  \\
ChatGLM3-6B & 40.0 & 75.4 & 41.7 & 82.4 & \ws{14.5}{25.5} & 79.4 & \ws{25.4}{14.6} & \ws{7.2}{32.8} & \ws{5.3}{34.7} & \ws{6.5}{33.5} \\
Mistral-7B & 49.4 & 53.9 & 96.2 & 57.9 & 73.9 & \ws{0.6}{48.8} & 75.9 & \ws{11.3}{38.1} & \ws{8.9}{40.5} & \ws{6.4}{43.0} \\
\midrule
\multicolumn{11}{c}{\textbf{few-shot ICL}} \\
\midrule
GPT-3.5-Turbo & 61.7 & 93.3 & \ws{23}{38.7} & 91.3 & \ws{38.4}{23.3} & 96.7 & \ws{52.4}{9.3} & \ws{11.7}{50.0} & \ws{13.9}{47.8} & \ws{12.2}{49.5} \\
GPT-4 & {83.9} & {96.7} & {99.3} & {91.3} & \ws{12.6}{71.3} & {94.0} & {92.0} & \ws{34.2}{49.7} & \ws{14.2}{69.7} & \ws{11.9}{72.0}\\
Gemini-Pro & {77.9} & {95.3} & {92.9} & {94.0} & \ws{40.9}{37.0} & {95.8} & \ws{47.5}{30.4} & \ws{32.3}{45.6} & \ws{22.6}{55.3} & \ws{17.4}{60.5} \\
LLaMA2-Chat-7B & {40.2} & {66.5} & {75.3} & {53.3} & {59.5} & {55.9} & \ws{7.1}{33.1} & \ws{3.2}{37.0} & \ws{0.8}{39.4} & \ws{1.6}{38.6} \\
LLaMA2-Chat-13B & {59.1} & {97.5} & \ws{10.6}{48.5} & {87.3} & \ws{46.7}{12.4} & {92.4} & \ws{47.0}{12.1} & \ws{8.8}{50.3} & \ws{5.1}{54.0} & \ws{5.8}{53.3} \\
LLaMA2-Chat-70B & 57.8 & 100.0 & \ws{54.2}{3.6} & 99.8 & \ws{54.7}{3.1}  & 99.6 & \ws{56.2}{1.6}  & \ws{12.6}{45.2} & \ws{4.2}{53.7} & \ws{7.0}{50.8}  \\
ChatGLM3-6B & {35.6} & {100.0} & \ws{35.6}{0.0} & {100.0} & \ws{35.6}{0.0} & {100.0} & \ws{35.6}{0.0} & \ws{3.1}{32.5} & \ws{3.0}{32.6} & \ws{0.9}{34.7} \\
Mistral-7B & {63.9} & {84.4} & {84.7} & {73.3} & \ws{6.2}{57.7} & {72.1} & \ws{15.9}{48.0} & \ws{23}{40.9} & \ws{16.3}{47.6} & \ws{7.5}{56.4} \\
\midrule
\multicolumn{11}{c}{\textbf{zero-shot CoT}} \\
\midrule
GPT-3.5-Turbo  & {64.7} & {75.3} & {77.3} & {65.3} & \ws{5.4}{59.3} & {78.7} & \ws{29.4}{35.3} & \ws{13.2}{51.5} & \ws{10.7}{54.0} &  \ws{4.0}{60.7} \\
GPT-4 & {81.3} & {94.0} & {100.0} & {98.0} & \ws{30.0}{61.3} & {96.0} & {94.0} & \ws{23}{58.3} & \ws{6.1}{75.2} & \ws{12}{69.3}\\
Gemini-Pro & {72.7} & \ws{4.7}{68.0} & {94.6} & \ws{6.8}{65.9} & \ws{16.4}{56.3} & {74.9} & \ws{13.8}{58.9} & \ws{7.5}{65.2} & \ws{14.5}{58.2} & \ws{12.7}{60.0} \\
LLaMA2-Chat-7B & {48.0} & {71.2} & \ws{2.0}{46.0} & {62.7} & \ws{5.9}{42.1} & {63.4} & \ws{13.9}{34.1} & \ws{4.2}{43.8} & \ws{2.5}{45.5} & \ws{0.5}{47.5}   \\
LLaMA2-Chat-13B & {56.3} & {59.7} & {74.6} & \ws{3.8}{52.5} & {56.8} & \ws{2.4}{53.9} & \ws{14.6}{41.7} & \ws{7.1}{49.2} & \ws{4.3}{52.0} & \ws{7.5}{48.8}  \\
LLaMA2-Chat-70B & {60.3} & {74.4} & {69.7} & {69.6} & \ws{15.6}{44.7} & {72.0} & \ws{35}{25.3} & \ws{3.7}{56.6} & \ws{6.6}{53.7} & \ws{10}{52.3} \\
ChatGLM3-6B & {48.9} & {82.9} & \ws{16.9}{32.0} & {81.4} & \ws{24.1}{24.8} & {76.0} & \ws{20.9}{28.0} & \ws{9.8}{39.1} & \ws{4.7}{44.2} & \ws{5.4}{43.5}\\
Mistral-7B & {69.6} & {76.5} & {94.7} & {83.7} & \ws{6.1}{63.5} & {71.2} & \ws{11.2}{58.4} & \ws{23.3}{46.3} & \ws{19.7}{49.9} & \ws{10.8}{58.8}  \\
\midrule
\multicolumn{11}{c}{\textbf{few-shot CoT}} \\
\midrule
GPT-3.5-Turbo  & {71.7} & {85.3} & {75.3} & {83.3} & \ws{16.4}{55.3} & {90.0} & \ws{49.7}{22.0} & \ws{18}{53.7} & \ws{11}{60.7} & \ws{8.7}{63.0} \\
GPT-4 & {83.0} & {95.3} & {100.0} & {94.7} & \ws{17.0}{66.0} & {95.3} & {88.0} & \ws{15.7}{67.3} & \ws{8.3}{74.7} & \ws{12.7}{70.3}\\
Gemini-Pro  & {72.4} & {86.1} & \ws{7.9}{64.5} & {81.4} & \ws{31.9}{40.5} & {87.5} & \ws{35.4}{37.0} & \ws{9.2}{63.2} & \ws{13.0}{59.4} & \ws{10.0}{62.4}  \\
LLaMA2-Chat-7B & {43.8} & {78.1} & \ws{8.9}{34.9} & {70.3} & \ws{6.1}{37.7} & {64.3} & \ws{1.7}{42.1} & \ws{4.5}{39.3} & \ws{2.4}{41.4} & \ws{3.0}{40.8} \\
LLaMA2-Chat-13B & {60.6} & {72.1} & \ws{9.5}{51.1} & \ws{6.1}{54.5} & \ws{23.4}{37.2} & {70.6} & \ws{28.0}{32.6} & \ws{13.1}{47.5} & \ws{10.0}{50.6} & \ws{7.5}{53.1}  \\
LLaMA2-Chat-70B & 70.9 & 78.2 & \ws{4.7}{66.2} & \ws{2.9}{68.0} & \ws{16.9}{54.0} &78.9 & \ws{32.5}{38.4} & \ws{12.4}{58.5} & \ws{13}{57.9} & \ws{13}{57.9}  \\
ChatGLM3-6B & {40.0} & {94.6} & \ws{30.3}{9.7} & {92.9} & \ws{28.6}{11.4} & {86.8} & \ws{20.0}{20.0} & \ws{5.2}{34.8} & \ws{5.3}{34.7} & \ws{1.3}{38.7}  \\
Mistral-7B & {67.6} & {88.3} & \ws{9.0}{58.6} & {84.0} & \ws{29.4}{38.2} & {81.9} & \ws{35.3}{32.3} & \ws{17.2}{50.4} & \ws{19.1}{48.5} & \ws{8.2}{59.4}  \\
\bottomrule
\end{tabular}
\caption{Accuracy ($\%$) across all datasets under four prompt settings. $E$ and $\neg E$ are respectively referring to entailment (IID) and non-entailment (OOD) sets. The intensity of \colorbox{cyan!50}{blue} highlights corresponds to the \emph{absolute} decrease in accuracy compared to the Standard dataset for each LLM.
}
\label{tab: acc}
}
\end{table*}

\textbf{Confidence Score (CFS)} is designed to evaluate LLMs' self-assessment capabilities. We follow  \cite{xiong2023can} to prompt LLMs to provide their confidence level, which indicates the degree of certainty they have about their answer and is represented as a percentage.

\subsection{Evaluated LLMs}
To obtain a comprehensive understanding of how LLMs are affected by shortcuts, we conduct experiments on three widely used closed-source LLMs: GPT-3.5-Turbo  \cite{OpenAI2023}, GPT-4  \cite{achiam2023gpt} and Gemini-Pro  \cite{team2023gemini}. Regarding open-source LLMs, we select LLaMA2-Chat-series (7B, 13B, 70B)  \cite{touvron2023llama}, ChatGLM3-6B  \cite{zeng2022glm} and Mistral-7B  \cite{jiang2023mistral} for assessment. 

\subsection{Prompting Strategies}
Our experiments aim to assess the performance of LLMs in different settings, including zero-shot, few-shot ICL, zero-shot CoT, and few-shot CoT promptings. For zero-shot CoT, we utilize the prompt depicted in Figure \ref{fig: example}. To construct few-shot ICL prompts, we enhance the best-performing zero-shot prompt by incorporating three random samples from the remaining examples in MultiNLI. Likewise, we employ a similar sampling approach for few-shot CoT and use GPT-4 to generate analyses for these examples.

\begin{figure*}[t]
\centering
\includegraphics[width=1\textwidth]{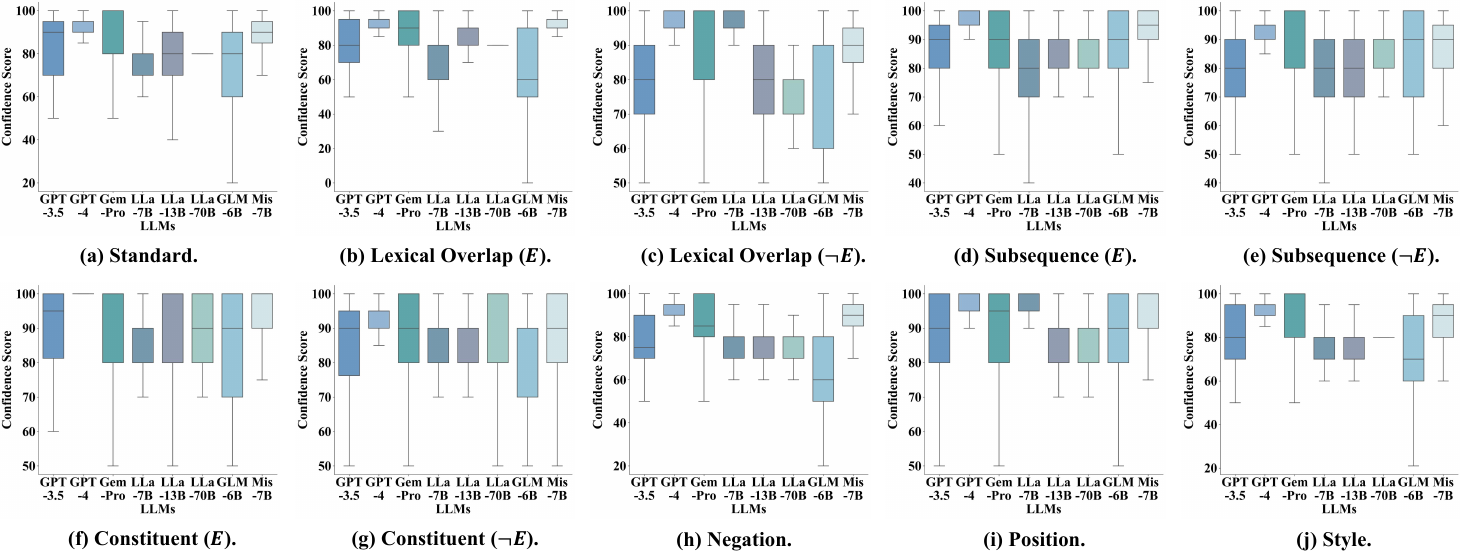}
\caption{Box plots of confidence scores across all datasets under zero-shot CoT prompting (each LLM is denoted by an abbreviation). LLMs generally report confidence scores that significantly exceed their actual accuracy.}
\label{fig: confidence}
\end{figure*}

\section{ Experimental Results }
We conduct our experiments based on the Shortcut Suite and observe that LLMs tend to exploit various shortcuts in downstream tasks, resulting in a notable decrease in performance. In this section, we present a comprehensive analysis.

\subsection{Overall Performance}
\subsubsection{Effect of Different LLMs}
As shown in Table \ref{tab: acc}, closed-source and some open-source LLMs excel on standard datasets, with GPT-4 leading at an accuracy of 85.6\%, followed by Gemini-Pro at 77.9\%, GPT-3.5-Turbo at 71.7\%, LLaMA2-Chat-70B at 70.9\% and Mistral-7B at 69.6\%. However, this high level of performance does not extend to datasets containing shortcuts. For example, the accuracy of GPT-3.5-Turbo on the Constituent ($\neg E$) dataset drops by 52.4\% in the few-shot ICL setting. This significant drop suggests that LLMs are easily prone to adopting shortcuts for prediction.

Among the open-source LLMs, Mistral-7B performs the best with CoT prompts. It excels on both standard and shortcut datasets, nearly surpassing LLaMA2-Chat-13B in all settings and even exceeding GPT-3.5-Turbo in some scenarios, demonstrating remarkable capabilities in NLI and robustness generalization. On the other hand, ChatGLM3-6B is the most affected by shortcuts, resulting in the poorest performance.

Furthermore, we observe an inverse scaling pattern of LLaMA2-Chat in zero-shot and few-shot ICL scenarios. As the model size increases, it tends to rely more on spurious mapping for NLI tasks, resulting in lower accuracy. However, in the CoT scenario, LLaMA2-Chat-70B outperforms smaller models on most datasets. This indicates that larger models retain improved semantic comprehension and reasoning abilities but require appropriate prompting to fully leverage their potential. This phenomenon is also observed in the LLaMA3 series, as illustrated in Appendix \ref{sec: appendixC}.

\subsubsection{Effect of Shortcut Types}
Regarding Lexical Overlap, Subsequence, and Constituent shortcuts, LLMs consistently favor predicting \textit{entailment} ($E$) and thus struggle with the \textit{non-entailment} ($\neg E$) class. This indicates that LLMs can easily exploit these spurious correlations with the label $E$, leading to poor performance on $\neg E$ instances. Lexical Overlap appears to be the easiest task for most LLMs across different prompt settings, resulting in consistently high accuracy, while the Constituent shortcut poses the greatest challenge. For instance, in the zero-shot setting, Gemini-Pro experiences a significant 29.0\% drop on Constituent, from 76.2\% to 47.2\%, worse than random guessing at 50\%.

Negation, Position, and Style shortcuts also prove challenging for most LLMs, as indicated by the notable decrease in accuracy. In the Negation dataset, the accuracy of GPT-4 decreases by 15-35\% across the four different prompt settings. In the Style dataset, the accuracy of GPT-4 decreases up to 15.6\%. Moreover, the detailed results of the Position shortcut are presented in Table \ref{tab:[position]}. The lowest accuracy rates are predominantly observed when extra phrases are added at the beginning of the sentence, suggesting that the LLMs may rely more heavily on the beginning parts of sentences for cues than the end parts, which could be a potential shortcut for improvement.

\begin{table}[t]
  \centering
\footnotesize{
\setlength{\tabcolsep}{4pt}
\begin{tabular}{lcccc}
\toprule
\multirow{1}{*}{\textbf{Model}} & \multicolumn{2}{c}{\textbf{premise}} & \multicolumn{2}{c}{\textbf{hypothesis}}  \\
  \cmidrule(lr){2-3}   \cmidrule(lr){4-5} 
& \textbf{start} & \textbf{end} & \textbf{start} & \textbf{end} \\
\midrule
GPT-3.5-Turbo & 61.3 & 56.0 & \uline{48.0} & 50.7 \\
GPT-4 & 77.6 & 79.7 & 76.4 & \uline{71.2}\\
Gemini-Pro & \uline{50.7} & 62.8 & 55.1 & 62.4 \\
LLaMA2-Chat-7B & 46.6 & 46.2 & \uline{42.1} &  46.3   \\
LLaMA2-Chat-13B & 50.0 & 57.9 & \uline{47.9} & 50.8  \\
LLaMA2-Chat-70B & \uline{51.8} & 62.0 & 53.8 & 55.1  \\
ChatGLM3-6B & 43.5 & 45.5 & \uline{42.1} & 44.1   \\
Mistral-7B & 49.7 & 50.6 & \uline{47.1} & 47.3  \\
\bottomrule
\end{tabular}
}
\caption{Accuracy Details for Position Shortcut: We place tautologies at the start or end of the premise or hypothesis in the Standard dataset. The lowest accuracy for each LLM is underlined, which frequently occurs when the tautologies are placed at the beginning of the source text. }
\label{tab:[position]}
\end{table}

\begin{figure*}[t]
\centering
\includegraphics[width=1\textwidth]{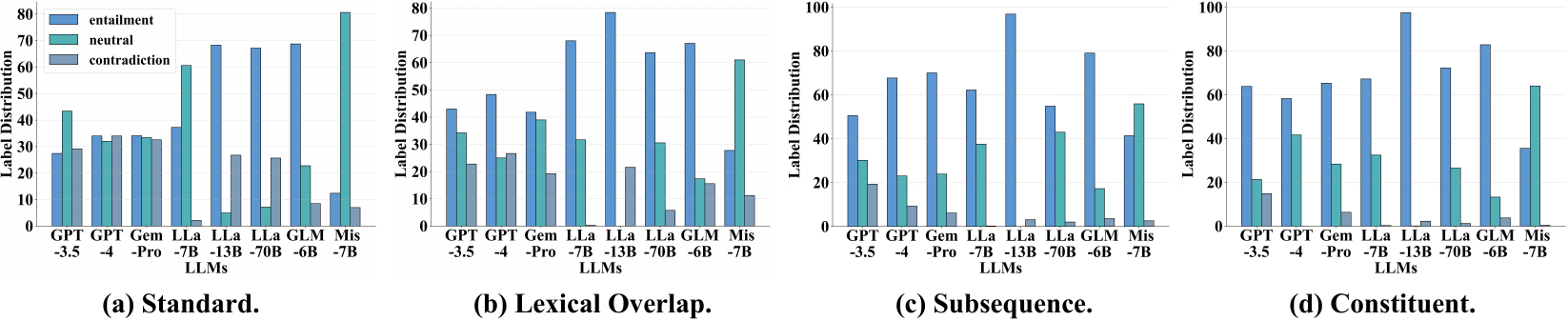}
\caption{Label distribution percentages (\%) for each LLM’s predictions under zero-shot prompting (each LLM is abbreviated). Distributions for the other three datasets are in Appendix \ref{sec: appendixA}.}
\label{fig: label}
\end{figure*}

\subsubsection{Effect of Prompting Types}
Most LLMs demonstrate significant performance gains in all datasets when utilizing the CoT prompt. For example, GPT-4 with a zero-shot CoT prompt on the Constituent ($\neg E$) dataset achieves an accuracy improvement of 14.0\% compared to zero-shot, while LLaMA2-Chat-13B shows an improvement of 40.9\% under the same conditions.  However, the accuracy of GPT-4 and Gemini-Pro decreases after applying the CoT prompt on the Standard dataset and Lexical Overlap dataset. This phenomenon reveals that LLMs are prone to utilize shortcuts to predict, and the CoT prompt can promote in-depth inference and reduce the reliance on spurious correlations, thus improving performance. However, for relatively simple datasets, advanced LLMs may already possess sufficient semantic understanding and reasoning capabilities, reducing their dependence on CoT for performance enhancement.

Additionally, it is worth noting that the effectiveness of few-shot prompts is not superior to zero-shot prompting. In several scenarios, the few-shot ICL is less effective than the zero-shot, and the few-shot CoT performs worse than the zero-shot CoT. This discrepancy could be attributed to the LLMs acquiring biases from the in-context examples. Similar phenomena have been reported in
 \cite{kim2023better, Tang_Kong_Huang_Xue_2023}. We show more experimental results and analysis in Appendix \ref{sec: appendixD}.

\subsection{In-depth Analysis}
\subsubsection{Explanation Quality}
We evaluate the explanation quality of LLMs in shortcut challenges using Equations \ref{eq: SFS}, \ref{eq: ICS}, and \ref{eq:eqs}, with results presented in Table \ref{tab: eqs}.

For SFS, most LLMs score above 85\%, indicating that current models have achieved a relatively high level of semantic fidelity. GPT-3.5-Turbo scores the highest on the Standard dataset with 92.1\%, while Mistral-7B scores the lowest at 88.5\%. Generally, models demonstrate a slight decline in SFS on shortcut datasets compared to the Standard dataset, indicating a reduced ability to restate inputs effectively in these contexts.

Regarding ICS, most LLMs score below 50\%, suggesting that more than half of their responses are contradictory. Notably, LLMs exhibit lower ICS scores on shortcut datasets compared to the Standard dataset. For example, LLaMA2-Chat-70B achieves a score of 41.5\% on the Standard dataset but only 13.5\% on the Negation dataset. These observations suggest that a lack of internal consistency in reasoning is a significant factor contributing to LLMs' reduced performance when dealing with shortcuts.

The overall EQS, which combines SFS and ICS, provides a comprehensive reflection of the overall quality of explanations from LLMs. Typically, models that exhibit higher accuracy also demonstrate greater explanatory capabilities.

\begin{table*}[t]
  \centering
  \resizebox{1\textwidth}{!}{
  
% 减小整个表格的字体大小
% \scriptsize{
% 减小列之间的间距
% \setlength{\tabcolsep}{4pt}
\begin{tabular}{lcccccccccccccc}
\toprule
\multirow{2}{*}{\textbf{Model}} & \multicolumn{1}{c}{\textbf{Standard}} & \multicolumn{2}{c}{\textbf{Lexical Overlap}} & \multicolumn{2}{c}{\textbf{Subsequence}} & \multicolumn{2}{c}{\textbf{Constituent}} & \multicolumn{1}{c}{\textbf{Negation}} & \multicolumn{1}{c}{\textbf{Position}} & \multicolumn{1}{c}{\textbf{Style}} \\
  \cmidrule(lr){3-4}   \cmidrule(lr){5-6}   \cmidrule(lr){7-8} 
& & \textbf{$E$} & \textbf{$\neg E$} & \textbf{$E$} & \textbf{$\neg E$} & \textbf{$E$} & \textbf{$\neg E$} & &  &   \\
\midrule
\multicolumn{11}{c}{\textbf{SFS | ICS}} \\
\midrule
GPT-3.5-Turbo & 92.1 | 29.0 & 91.0 | 35.3  & 92.0 | \uline{5.3} & 91.0 | 30.5  & 91.5 | 25.7 & \uline{89.5} | 36.6  & 90.8 | 26.1 & 93.3 | 21.7 & 92.5 | 25.7 & 92.3 | 22.7 \\
GPT-4 & 91.1 | 34.7 & 91.1 | 35.3 & 91.3 | \uline{11.3} & 90.8 | 23.3 & 90.0 | 23.3 & 91.8 | 42.7 & 89.2 | 18.0 & \uline{88.7} | 57.0 & 91.8 | 43.7 & 90.2 | 28.3 \\
Gemini-Pro & 89.2 | 43.0 & 88.6 | 39.0  & 88.4 | 29.9 & 87.9 | 30.5 & 88.8 | \uline{25.7} & \uline{87.3} | 36.6 & 90.0 | 26.1 & 90.8 | 46.4 &  89.2 | 40.0 & 89.1 | 47.7 \\
LLaMA2-Chat-7B & 88.7 | 20.3 & 90.6 | 29.5 & 90.1 | \uline{4.1} & 90.2 | 24.2 & 90.4 | 15.8 & 90.8 | 23.0 & 90.4 | 16.7 & 89.8 | 11.1 & 90.1 | 15.2 & \uline{88.6} | 19.8 \\
LLaMA2-Chat-13B & 90.2 | 41.5 & 91.4 | 31.2 & 91.1 | \uline{11.0}& 91.3 | 26.5 & 90.0 | 23.5 & 92.3 | 36.4 & 90.8 | 25.5 & \uline{88.4} | 13.5 & 92.3 | 18.0 & 89.9 | 25.0\\
LLaMA2-Chat-70B & 90.4 | 33.9 & 90.6 | 42.1 & 91.1 |\uline{ 6.9 }& 90.1 | 36.7 & 90.5 | 24.0 & 90.3 | 41.9 & 90.4 | 34.0 & 90.3 | 25.4 & 91.3| 30.9 & \uline{90.0} | 30.4 \\
ChatGLM3-6B & 90.3 | 22.9 & 87.7 | 24.5 & 88.1 | \uline{9.5} & 88.0 | 22.4 & 88.0 | 21.2 & 87.8 | 20.1 & \uline{87.7} | 24.0& 91.2 | 24.2 & 90.5 | 23.3 & 90.4 | 23.5 \\
Mistral-7B & 88.5 | 45.5 & 85.1 | 63.9 & 89.0 | \uline{29.4} & 84.2 | 67.7 & 88.3 | 54.9 & \uline{83.2} | 69.2 &  87.9  | 53.0 & 91.2 | 44.4 & 87.2 | 49.6  & 89.5 | 44.2 \\
\midrule
\multicolumn{11}{c}{\textbf{EQS}} \\
\midrule
GPT-3.5-Turbo & 60.6 & 63.2 & \uline{48.7} & 60.8  & 58.6 &  63.1 & 58.5 & 57.5 & 59.1 & 57.5 \\
GPT-4 & 62.9 & 63.2 & \uline{51.3} &  57.1 & 56.7 & 67.3 & 53.6 & 72.9 & 67.8 & 59.3 \\
Gemini-Pro & 66.1 & 63.8 & 59.2 & 59.2 & \uline{57.3} &  62.0 &  58.1 & 68.6 & 64.6 & 68.4 \\
LLaMA2-Chat-7B & 54.5 & 60.1 & \uline{47.1} & 57.2  & 53.1 & 56.9 & 53.6 & 50.5 & 52.7 & 54.2 \\
LLaMA2-Chat-13B & 65.9 & 61.3 & 51.1 & 58.9 & 56.8 & 64.4 & 58.2 & \uline{51.0} & 55.2 & 57.5 \\
LLaMA2-Chat-70B & 62.2 & 66.4 & \uline{49.0} & 126.8 & 57.3 & 66.1 & 62.2 & 57.9 & 61.1 & 60.2  \\
ChatGLM3-6B & 56.6 & 56.1 & 48.8 & \uline{55.2} & 54.6 &  54.0 & 55.9 & 57.7 & 56.9 & 57.0 \\
Mistral-7B & 67.0 & 74.5 & \uline{59.2} & 76.0 & 71.6 & 76.2 & 70.5 & 67.8 & 68.4 & 66.9 \\
\bottomrule
\end{tabular}
}
\caption{SFS (\%), ICS (\%), and EQS (\%) across all datasets under zero-shot CoT prompting. The worst score for each LLM is underlined. LLMs typically show the lowest explanation quality in datasets comprising shortcuts.}
\label{tab: eqs}
\end{table*}

\subsubsection{Confidence Score}

Figure \ref{fig: confidence} displays the confidence levels of LLMs, revealing two key findings. First, LLMs tend to be overconfident, with their confidence scores rarely falling below 60\% and often significantly exceeding their actual accuracy. Second, the discrepancy between confidence and accuracy is notably greater in datasets containing shortcuts compared to the Standard dataset. This suggests that LLMs not only adopt shortcuts but also exhibit heightened confidence in these spurious mappings without fully understanding the true relationship between the source text and the corresponding label.

\subsubsection{Prediction Distribution}
Figure \ref{fig: label} shows the label distribution in each LLM's prediction.  Despite a balanced distribution in the ground truth, we can easily observe that in the Standard dataset, GPT-3.5-Turbo, LLaMA2-Chat-7B, and Mistral-7B tend to disproportionately predict \textit{neutral} over the other two categories. Conversely, LLaMA2-Chat-13B and ChatGLM3-6B show a bias towards \textit{entailment}. This pattern may stem from multiple factors, including potential overfitting to the NLI task or tasks with a similar categorical structure. 

For datasets featuring Lexical Overlap, Subsequence, and Constituent shortcuts, LLMs predominantly predict \textit{entailment}, indicating a susceptibility to these shortcuts. For the Negation shortcut, a rise in \textit{contradiction} predictions by GPT-4 and LLaMA2-Chat-13B suggests a reliance on a spurious correlation between negation words and the \textit{contradiction} label. 

\subsubsection{Error Analysis}
\begin{figure}[t]
\centering
\includegraphics[width=0.48\textwidth]{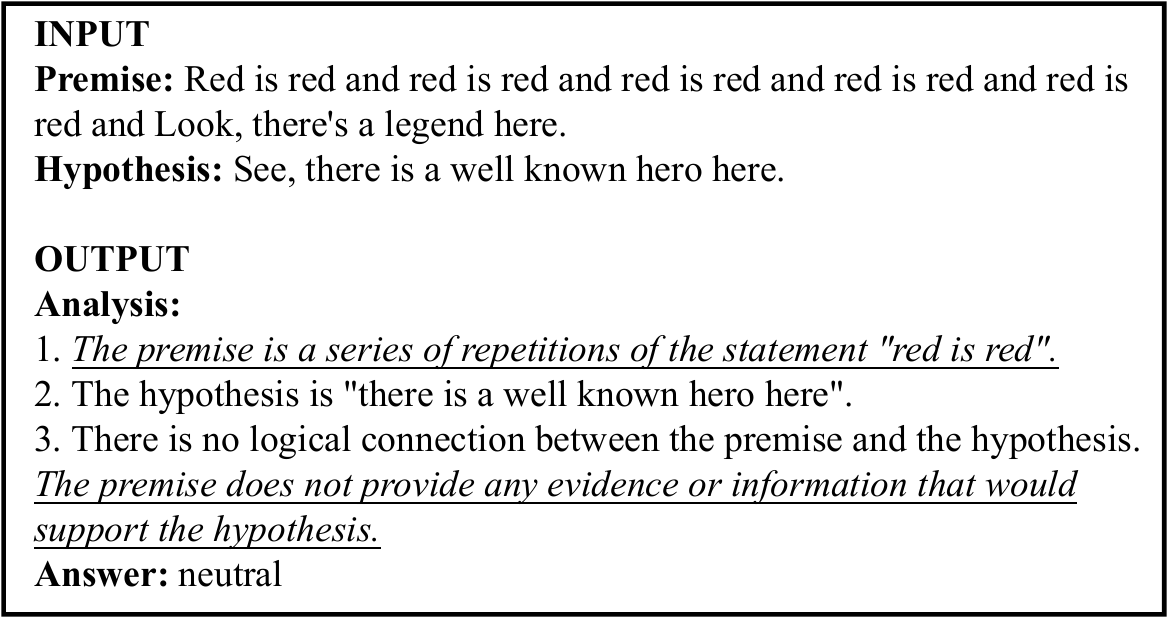}
\caption{An illustrative example of distraction in LLMs: in the Position dataset, the LLM is observed to be distracted by tautologies, thus ignoring useful information.}
\label{fig: distracting}
\end{figure}

We identify three types of errors in shortcut learning by analyzing the CoT responses of LLMs. The first issue is \textbf{distraction}, where LLMs are easily distracted by irrelevant information. As shown in Figure \ref{fig: distracting}, they may focus on repetitive tautologies, leading to the neglect of useful information in the original text.  Additionally, they often prioritize words at the start of a sentence while neglecting those at the end, as shown in Table \ref{tab:[position]}.
This reflects a tendency in LLMs to concentrate on local information while ignoring the comprehensive context. 

Second, LLMs suffer from \textbf{disguised comprehension}.  Specifically, they struggle to grasp the subtleties of individual words, sentence structures, and complex biblical language styles, shifting one's concept to another. This leads to disguised comprehension where LLMs might inadvertently ``borrow'' concepts, causing them to rely on shortcuts to make incorrect inferences. The detailed case can be found in Figure \ref{fig: disguised}.

The third issue is \textbf{logical fallacy}. LLMs tend to reduce intricate reasoning to overly simplistic terms, generalizing from specific instances to broader conclusions  via the use of shortcuts. This oversimplification in their reasoning process can lead to erroneous results, as illustrated in Figure \ref{fig: fallacy}.

\subsection{Extended Evaluation}
To gain further insight into the shortcut challenges in LLMs, we  conduct experiments on other NLP tasks. The first is the Sentiment Analysis (SA) task. Specifically, we use the validation set of the Stanford Sentiment Treebank (SST-2)  \cite{socher2013recursive} as our Standard dataset. We then introduce the Negation shortcut using the method described in Section \ref{dataset} to the Standard dataset.  
The second is the Paraphrase Identification (PI) task. We experiment with the Quora Question Pairs (QQP) \footnote{The dataset is available at \url {https://www.kaggle.com/c/quora-question-pairs}.} dataset as Standard dataset and the Paraphrase Adversaries from Word Scrambling (PAWS)  \cite{zhang2019paws}  dataset to represent Lexical Overlap shortcut. The results, presented in Table \ref{tab:sa}, demonstrate a consistent decline in performance across both the SA and PI tasks on datasets comprising shortcuts compared to Standard datasets. Furthermore, as shown in Figure \ref{fig: sst2}, there is a noticeable increase in \textit{negative} predictions on the Negation dataset and an increase in \textit{duplicate} predictions on the Lexical Overlap dataset. This pattern suggests that LLMs tend to capture spurious correlations between negation words and the \textit{negative} label, as well as between word overlap and the \textit{duplicate} label. In conclusion, we find that LLMs are prone to relying on the Negation shortcut in the SA task and the Lexical Overlap shortcut in the PI task, suggesting that shortcut learning is a prevalent phenomenon in LLMs across a wide spectrum of tasks.

Besides the LLMs mentioned above, we also conduct experiments on the latest LLMs, such as LLaMA3-series, and analyze the results as detailed in Appendix \ref{sec: appendixC}.

\begin{table}[!t]
  \centering
\small{
\setlength{\tabcolsep}{0.4pt}
\begin{tabular}{lccccc}
\toprule
\multirow{1}{*}{\textbf{Model}} & \multicolumn{2}{c}{\textbf{SA}} & \multicolumn{2}{c}{\textbf{PI}}  \\
  \cmidrule(lr){2-3}   \cmidrule(lr){4-5}  
 & \textbf{Standard} & \textbf{Negation} & \textbf{Standard} & \textbf{Overlap}   \\
\midrule
GPT-3.5-Turbo & 91.7 & \ws{4.7}{87.0} & 81.2 & \ws{6.9}{74.3}  \\
GPT-4 & 93.0 & \ws{2.8}{90.2} & 73.7 & \ws{9.5}{64.2} \\
Gemini-Pro & 92.7 & \ws{4.9}{87.8} & 75.9 & \ws{28.5}{47.4}\\
LLaMA2-Chat-7B & 84.1 & \ws{8.0}{76.1} & 61.6 & \ws{12.1}{49.5}\\
LLaMA2-Chat-13B & 87.4 & \ws{4.1}{83.3} & 73.8 & \ws{23.8}{50.0}\\
LLaMA2-Chat-70B & 87.8 & \ws{0.7}{87.1} & 71.7 & \ws{19.7}{52.0} \\
ChatGLM3-6B & 90.4 & \ws{5.0}{85.4} & 64.9 & \ws{15.3}{49.6} \\
Mistral-7B & 80.5 & \ws{1.4}{79.1} & 52.6 & \ws{3}{49.6}\\
\bottomrule
\end{tabular}
}
\caption{Accuracy (\%) of the SA and PI tasks under zero-shot prompting. LLMs consistently demonstrate reduced performance on shortcut datasets compared to the Standard, as indicated by the \colorbox{cyan!50}{blue} highlights.}
\label{tab:sa}
\end{table}

\section{Conclusion}
\label{sec:conclusion}
In this study, we proposed Shortcut Suite, a test suite designed to evaluate the performance of LLMs in shortcut learning across several NLP tasks. Shortcut Suite encompasses six types of shortcuts: Lexical Overlap, Subsequence, Constituent, Negation, Position, and Style, and evaluates performance using five metrics: ACC, SFS, ICS, EQS, and CFS, across four prompt settings: zero-shot, few-shot ICL, zero-shot CoT, and few-shot CoT. 
Our extensive experiments on diverse LLMs demonstrated that LLMs frequently rely on shortcuts in downstream tasks. We explored the impact of different models, types of shortcuts, and prompting strategies. Our analysis then extended to explanation quality, label distribution, confidence score and error analysis.

Our findings offer new perspectives on LLMs' robustness and present new challenges for reducing their shortcut reliance, paving the way for future advancements in this field.

\section{Limitations}

In this paper, we primarily focus on evaluating the effect of shortcut learning in LLMs on the NLI task, with additional exploration into tasks like SA and PI. However, we acknowledge that other NLP tasks, such as question-answering and coreference inference, could offer further insights and should be investigated in future research. 

While this study provides a comprehensive understanding of shortcut learning in LLMs, it does not propose specific methods to mitigate this phenomenon effectively. Nonetheless, we identify shortcut learning behavior in LLMs and categorize potential error types associated with shortcut learning, offering a foundation for future research.
Based on our findings, we suggest several potential approaches for addressing shortcut learning in LLMs. One approach is fine-tuning on unbiased datasets, as training models on diverse and representative datasets may help alleviate shortcut learning. Moreover, employing advanced prompting techniques is essential. Our experiments indicate that few-shot prompting is insufficient for mitigating shortcut learning behaviors in LLMs, thus enhancing reasoning capabilities through methods such as CoT prompting may prove effective. Additionally, implementing retrieval augmentation by incorporating relevant external documents can ground LLMs, thereby reducing knowledge gaps and instances of hallucination. We advocate for further research to develop effective strategies aimed at addressing shortcut learning in LLMs.

\section*{Acknowledgments}
This research was partially supported by grants from the Joint Research Project of the Science and Technology Innovation Community in Yangtze River Delta (No. 2023CSJZN0200), the National Natural Science Foundation of China (No. 62337001), Anhui Provincial Natural Science Foundation (No. 2308085QF229) and the Fundamental Research Funds for the Central Universities (No. WK2150110034).

\newpage
% Entries for the entire Anthology, followed by custom entries
\bibliography{acl2024}
\bibliographystyle{acl_natbib}
\clearpage
\newpage
\appendix

\section{Appendix: Label Distribution}
\label{sec: appendixA}
% \captionsetup{width=1\textwidth}
\begin{figure}[h]
% \centering
\includegraphics[width=1.0\textwidth]{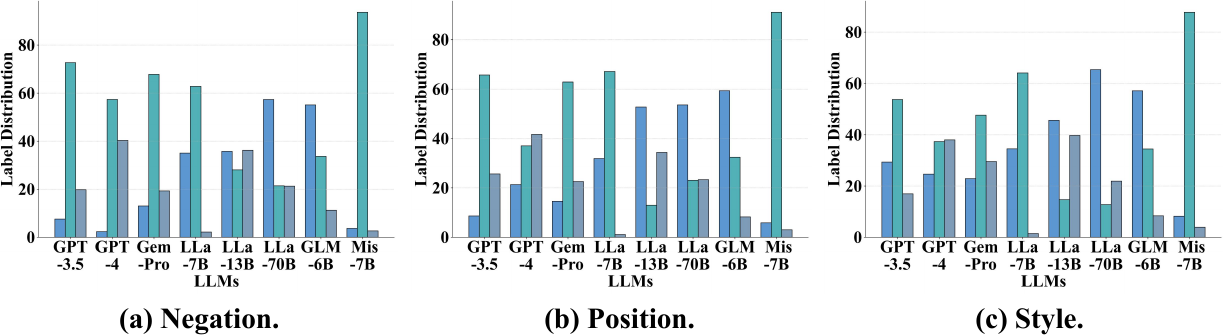}
\captionsetup{width=1\textwidth, justification=raggedright}
% \hspace{-10cm}  
\begin{minipage}[t]{1\textwidth}  % 设置 caption 的宽度
    \caption{Label distribution as percentages (\%) for LLMs' prediction under zero-shot prompting (each LLM is denoted by an abbreviation).}
    \label{fig: label2}
\end{minipage}
% \caption{Label distribution as percentages (\%) for LLMs' prediction under zero-shot prompting. Each LLM is denoted by an abbreviation. }
\label{fig: label2}
\end{figure}
\vspace*{-0.5cm}
% \twocolumn
\section{Appendix: Error Analysis}
\label{sec: appendixB}
\begin{figure}[h]
\centering
\includegraphics[width=0.48\textwidth]{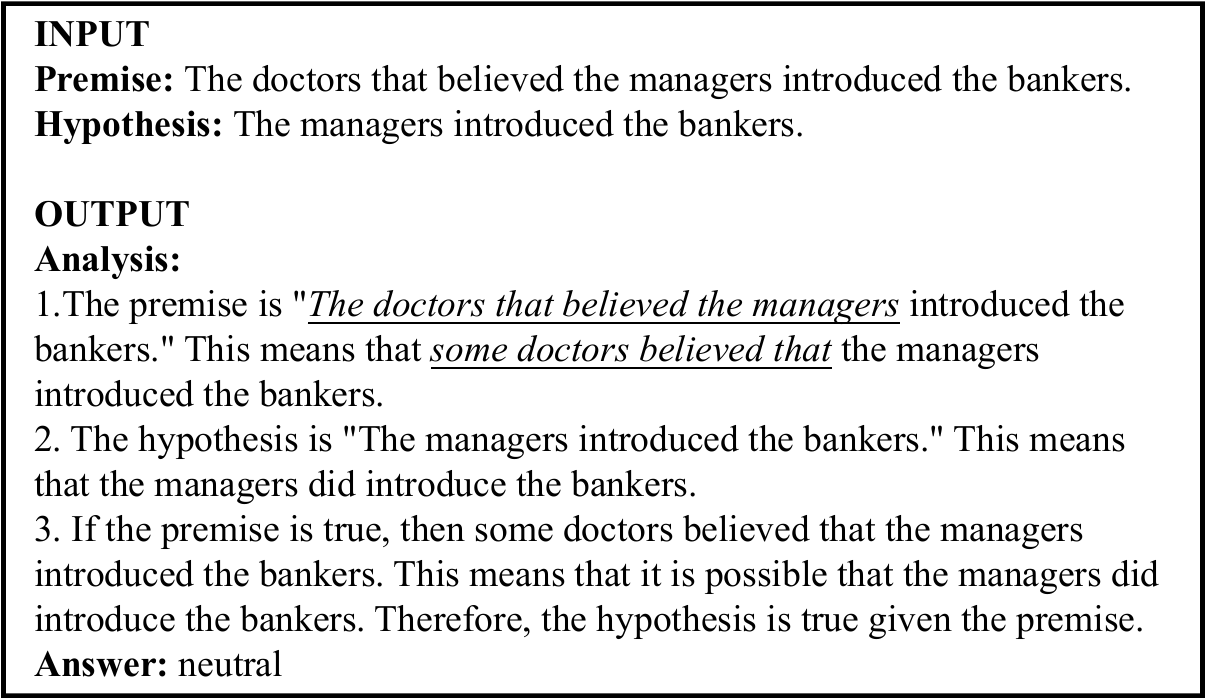}
\caption{An illustrative example of disguised comprehension in LLMs: the positions of ``that'' and ``believed'' are mistakenly swapped, leading to the incorrect assumption that they convey the same meaning.}
\label{fig: disguised}
\end{figure}

\begin{figure}[!h]
\centering
\includegraphics[width=0.48\textwidth]{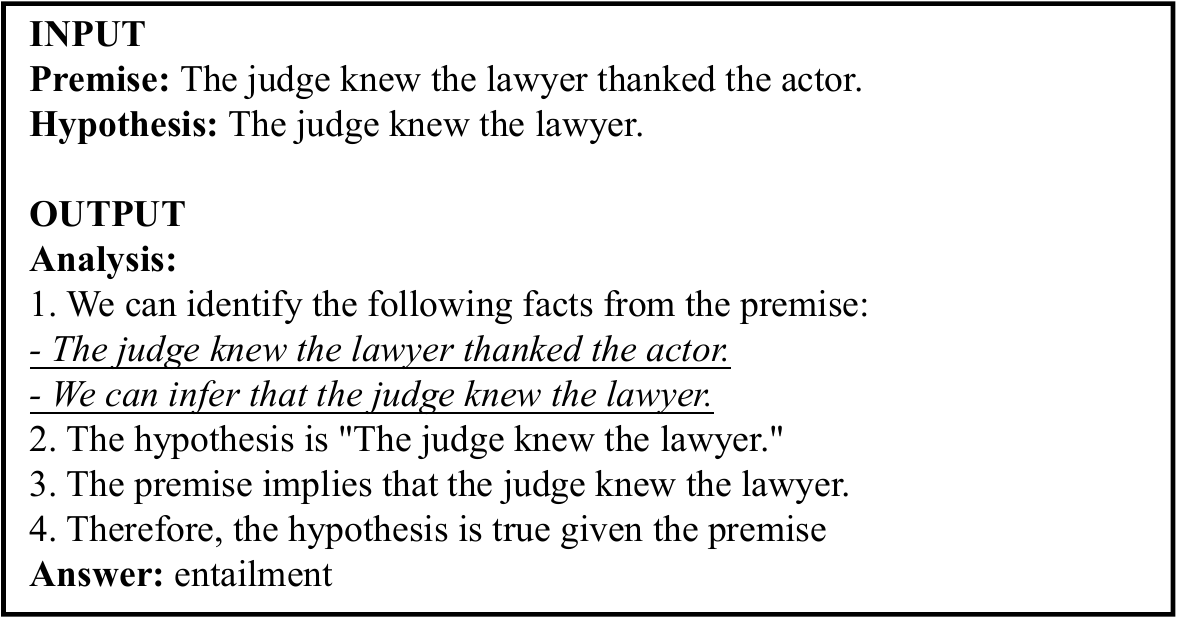}
\caption{ An illustrative example of logical fallacy in LLMs: an oversimplification in the Subsequence dataset is found in the analysis process. In the source text, knowing of an action (the lawyer thanking the actor) doesn't necessarily equate to knowing the person (the lawyer) in a broader sense. }
\label{fig: fallacy}
\end{figure}

Figure \ref{fig: disguised} and \ref{fig: fallacy} show the disguised comprehension error example and the logical fallacy error example respectively.

\vspace*{6.75cm}
\section{Appendix: Extended Evaluation of Shortcut Learning} 
\label{sec: appendixC}
% \vspace{7cm}
\paragraph{Model.}
In addition to the LLMs we discussed above, we'd like to extend our investigation to the LLaMA3-series. Notably, LLaMA3 demonstrates superior performance over LLaMA2. Specifically, LLaMA3-8B-Instruct outperforms both LLaMA2-Chat-7B and LLaMA2-Chat-13B on most datasets. Furthermore, LLaMA3-70B-Instruct surpasses GPT-3.5-Turbo and approaches the performance of Gemini-Pro. Despite these advances, we observe a consistent decline in performance on shortcut datasets compared to standard datasets. This trend suggests that LLaMA3-8B, similar to its predecessor, may rely on shortcuts for predictions. Additionally, the reverse scaling pattern persists in shortcut datasets such as Subsequence ($\neg E$) and Constituent ($\neg E$).
These supplementary experiments highlight the propensity of most LLMs to rely on shortcuts across a wide spectrum of tasks, underscoring the need for more robust and generalizable mechanisms.

\begin{figure*}[h]
\centering
\includegraphics[width=0.75\textwidth]{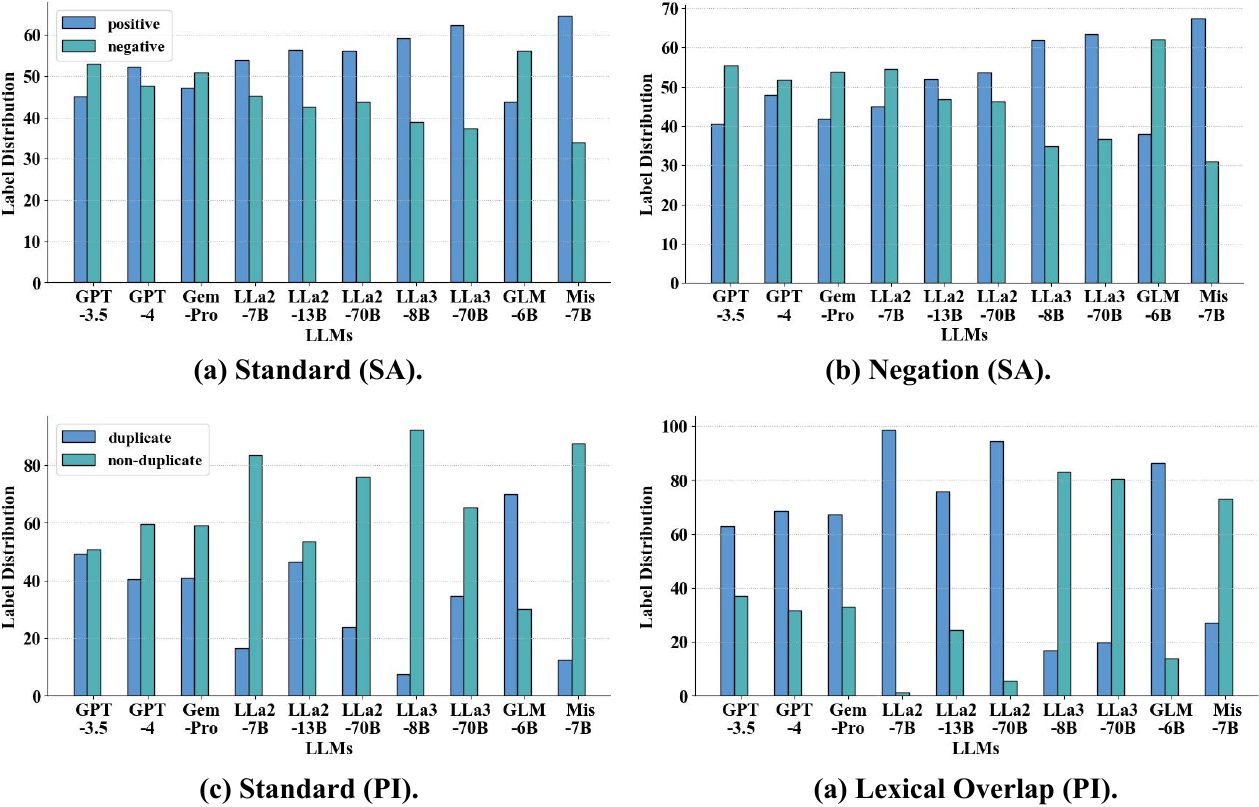}
\caption{ Label distribution as percentages (\%) for LLMs' prediction under zero-shot prompting on SA and PI task ( each LLM is denoted by an abbreviation). }
\label{fig: sst2}
\end{figure*}

\begin{table*}[!h]
    \centering
\small{
\setlength{\tabcolsep}{4pt}
\begin{tabular}{lcccccccccccccc}
\toprule
\multirow{2}{*}{\textbf{Model}} & \multicolumn{1}{c}{\textbf{Standard}} & \multicolumn{2}{c}{\textbf{Lexical Overlap}} & \multicolumn{2}{c}{\textbf{Subsequence}} & \multicolumn{2}{c}{\textbf{Constituent}} & \multicolumn{1}{c}{\textbf{Negation}} & \multicolumn{1}{c}{\textbf{Position}} & \multicolumn{1}{c}{\textbf{Style}} \\
  \cmidrule(lr){3-4}   \cmidrule(lr){5-6}   \cmidrule(lr){7-8} 
& & \textbf{$E$} & \textbf{$\neg E$} & \textbf{$E$} & \textbf{$\neg E$} & \textbf{$E$} & \textbf{$\neg E$} & &  &   \\
\midrule
\multicolumn{11}{c}{\textbf{zero-shot}} \\
\midrule
LLaMA3-8B-Instruct & 62.2 & 84.3 & 89.2 & 88.3 & \ws{13.9}{48.3} & 79.0 & \ws{22.1}{40.1} & \ws{10.6}{51.6} & \ws{9}{53.2} & \ws{7.2}{55.0} \\
LLaMA3-70B-Instruct & 74.5 & 94.3 & 96.8 & 99.7 & \ws{34.6}{39.9} & 83.9 & \ws{63.4}{11.1} & \ws{14.8}{59.7} & \ws{10.8}{63.7} & \ws{10.5}{64.0} \\
\midrule
\multicolumn{11}{c}{\textbf{zero-shot CoT}} \\
\midrule
LLaMA3-8B-Instruct & 65.3 & 63.5 & 96.1 & \ws{18.4}{46.9} & 75.7 & 65.3 & 68.6 & \ws{12.9}{52.4} & \ws{8.3}{57.0} & \ws{9.4}{55.9} \\
LLaMA3-70B-Instruct & 79.0 & 79.2 & 99.1 & 93.9 & \ws{20.8}{58.2} & \ws{30.5}{48.5} & \ws{7.4}{71.6} & \ws{16.9}{62.1} & \ws{13.6}{65.4} & \ws{27.3}{51.7} \\
\bottomrule
\end{tabular}
\caption{Accuracy ($\%$) across all datasets of LLaMA3-series. 
}
}
\label{tab: llama3}
\end{table*}

\begin{table*}[!h]
    \centering
\small{
\setlength{\tabcolsep}{4pt}
\begin{tabular}{lcccccccccccccc}
\toprule
\multirow{2}{*}{\textbf{Prompting}} & \multicolumn{1}{c}{\textbf{Standard}} & \multicolumn{2}{c}{\textbf{Lexical Overlap}} & \multicolumn{2}{c}{\textbf{Subsequence}} & \multicolumn{2}{c}{\textbf{Constituent}} & \multicolumn{1}{c}{\textbf{Negation}} & \multicolumn{1}{c}{\textbf{Position}} & \multicolumn{1}{c}{\textbf{Style}} \\
  \cmidrule(lr){3-4}   \cmidrule(lr){5-6}   \cmidrule(lr){7-8} 
& & \textbf{$E$} & \textbf{$\neg E$} & \textbf{$E$} & \textbf{$\neg E$} & \textbf{$E$} & \textbf{$\neg E$} & &  &   \\
\midrule
zero-shot & 56.7 & 69.5 & 83.8 & 58.6 & 58.3 & 67.5 & \ws{16.5}{40.2} & \ws{16.9}{39.8} & \ws{13.4}{43.3} & \ws{5.2}{51.5} \\
few-shot (MNLI)	 & 61.7 & 93.3 & \ws{23}{38.7} & 91.3 & \ws{38.4}{23.3} & 96.7 & \ws{52.4}{9.3} & \ws{11.7}{50.0} & \ws{13.9}{47.8} & \ws{12.2}{49.5} \\
few-shot (shortcut) & 61.7 & 86.3 & 90.3 & 81.7 & \ws{5.4}{56.3} & 82.3 & \ws{26.7}{35.0} & \ws{15.7}{46.0} & \ws{7.1}{54.6} & \ws{6.0}{55.7} \\
\bottomrule
\end{tabular}
\caption{Accuracy ($\%$) across all datasets of GPT-3.5-Turbo.}
\label{tab: few-shot}
}
\end{table*}

\section{Appendix: More Discussion on Few-shot Prompting}
\label{sec: appendixD}

As discussed above, few-shot ICL is less effective than zero-shot prompting, and few-shot CoT performs worse than zero-shot CoT in several scenarios. This phenomenon may be due to biases introduced by the in-context examples used in few-shot prompting. Similar issues have been reported in other studies. For instance,   \citet{kim2023better} observed that demonstrations can introduce biases, leading to reduced performance in language models.   \citet{Tang_Kong_Huang_Xue_2023} also noted that LLMs might exploit shortcuts in in-context learning, resulting in sub-optimal performance. Moreover, some papers focus specifically on this issue. For instance,   \citet{min2022rethinking} found that factors like the label space, the distribution of the input text, and the overall format of the sequence are critical determinants of task performance.
To further explore this issue, we conducted additional experiments using random samples from the remaining examples in each shortcut-laden dataset, beyond those from the MultiNLI dataset initially used in above experiments. The detailed results are shown in Table \ref{tab: few-shot}. We observe that LLMs' performance on shortcut-laden datasets using more similar examples is better than using standard examples, but still worse than zero-shot, indicating that the influence of shortcuts from pre-trained data is more significant than the benefits of in-context examples. LLMs struggle to summarize the important aspects from in-context examples to overcome their inherent biases and are even influenced by the biases from the in-context examples.

\end{document}